\documentclass{article} 
\usepackage{iclr2027_conference,times}
\makeatletter
\def\paragraph{\@startsection {paragraph}{4}{\z@}{0.8ex plus
0.3ex minus .2ex}{-1em}{\normalsize\bf}}
\makeatother

\usepackage{chngcntr}

\usepackage{amsmath,amsfonts,bm}

\def\eqref#1{equation~\ref{#1}}

\def\1{\bm{1}}

\DeclareMathAlphabet{\mathsfit}{\encodingdefault}{\sfdefault}{m}{sl}
\SetMathAlphabet{\mathsfit}{bold}{\encodingdefault}{\sfdefault}{bx}{n}

\usepackage{hyperref}
\usepackage{url}

\usepackage{booktabs}
\usepackage{amsfonts}
\usepackage{comment}
\usepackage{amsmath}
\usepackage{graphicx}
\usepackage{multirow}

\usepackage{xspace}
\newcommand{\method}{NUViT\xspace}
\newcommand{\methods}{NUViTs\xspace}

\title{NUViT: Vision Transformers for Fourier-Space Data}

\author{%
  Jeffrey Gu \\
  Princeton University\\
  \texttt{jg1427@princeton.edu} \\
  \And
  Robert C. Heeter \\
  Princeton University\\
  \texttt{robert.heeter@princeton.edu} \\
  \And
  Ellen D.~Zhong \\
  Princeton University \\
  \texttt{zhonge@princeton.edu} \\
}

\iclrfinalcopy 
\begin{document}

\maketitle

\begin{abstract}
Vision Transformers (ViTs) have demonstrated strong performance in many computer vision tasks but do not naturally extend beyond the natural image domain. One important example is Fourier-space data, which is used to represent a rich and diverse set of modalities, including medical imaging and microscopy. Standard ViT tokenization partitions an image into square patches on a Cartesian grid, which poorly fit the frequency and phase structure of Fourier-space data. We introduce \textbf{N}on-\textbf{U}niform \textbf{ViT}, a simple, general-purpose vision transformer that redesigns image tokenization, data representation, and positional encoding to respect this structure. Experiments on Fourier-domain image classification, 2D crystal lattice identification, MRI annotation, and cryo-EM protein heterogeneity analysis show that NUViT consistently captures the structure of Fourier data better than standard architectures.

\end{abstract}

\section{Introduction}

Vision Transformers (ViTs)~\citep{dosovitskiy2020image} have become a widely-used architecture across a broad range of computer vision tasks. Motivated by their strong performance on natural images, there is growing interest in extending ViT-style models to other data modalities. However, Fourier-domain or Fourier-transformed representations are not inherently compatible with standard ViT designs, even though Fourier data is common in applications such as 3D sensing, medical imaging, and microscopy.


General-purpose architectures for Fourier-space data remain underexplored, and existing frequency-space ViTs are largely domain-specific~\citep{buchholz2022fourier, duan2022fourier, alshaabi2025fourier}. Such a general transformer architecture would be valuable for 3D reconstruction methods that exploit the Fourier Slice Theorem~\citep{bracewell1956strip} to avoid costly real-space integration, a standard practice in cryo-electron microscopy (cryo-EM), and in experimental modalities where sensor data is collected natively in frequency space, such as magnetic resonance imaging (MRI). It would also benefit hybrid methods that incorporate Fourier transforms into their processing pipeline, including a growing body of work on frequency-space token mixing as an alternative to attention~\citep{li2020fourier, rao2021global, guibas2021adaptive, lee2022fnet, nguyen2022fourierformer, lingsch2023beyond, long2025stft}.

\begin{figure}
    \centering
    \includegraphics[width=\linewidth]{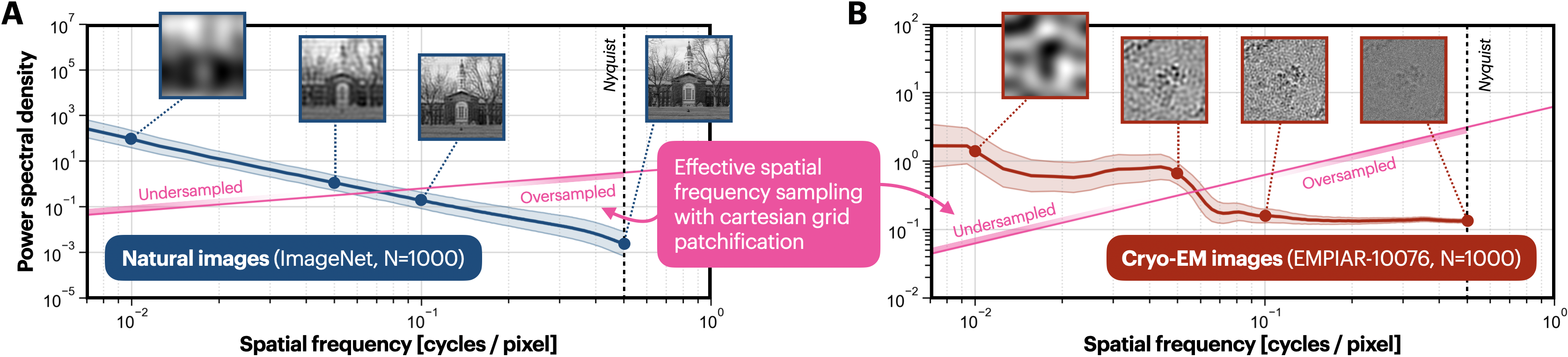}
    \vspace{-12pt}
    \caption{\textbf{Power spectral density of natural and cryo-EM images.} Log-log plots of radially-averaged power versus spatial frequency for \textbf{(A)} natural images~\citep{deng2009imagenet} and \textbf{(B)} cryo-EM images~\citep{davis2016modular} demonstrate a characteristic Fourier power-law decay. Shaded regions indicate one standard deviation. Pink lines show the spatial frequency weighting induced by Cartesian grid patching in standard ViTs, which undersamples informative low-frequency components. Insets show low-pass filtered images at different spatial frequencies. In both natural and cryo-EM images, lower spatial frequencies capture broad semantic information. Note that cryo-EM images inherently contain imaging aberrations (CTF), producing characteristic power oscillations.}
    \label{fig:motivation}
    \vspace{-6pt}
\end{figure}

Natural phenomena display a wide range of behavior in Fourier space. For example, speech and audio waveforms, natural images and video~\citep{field1987relations, van1996modelling}, 3D sensor measurements, and medical imaging such as MRI~\citep{lustig2007sparse} all follow a $1/f^\alpha$ spectral energy decay with spatial frequency $f$ (Figure~\ref{fig:motivation}). This Fourier power law is not the only common example of frequency-dependent behavior: also common are resonance- and dispersion-based phenomena. Resonance produces sharp peaks in a signal's Fourier spectrum, which is observed in atomic spectra and crystal structures. Dispersion describes how a wave's propagation speed varies with frequency, causing phenomena such as a prism splitting white light into its constituent colors.

In this paper, we design a general-purpose transformer architecture for Fourier-space data that better captures these patterns. We observe that the standard ViT tokenization strategy that partitions images into square patches on a uniform Cartesian grid is ill-suited for capturing the wide variety of structures present in Fourier-space data. We argue that the non-uniform frequency and phase coordinates of the Cartesian grid hamper the encoder's ability to understand Fourier space structure, with uniform frequency and phase coordinates providing better inductive bias. Furthermore, in the common case that a signal follows the Fourier power law, traditional ViT tokenization oversamples high spatial frequencies and undersamples the information-dense low frequency components. 

Based on this insight, we propose \textbf{N}on-\textbf{U}niform \textbf{Vi}sion \textbf{T}ransformers (NUViTs), a general-purpose transformer architecture for Fourier data. \methods sample the Fourier transform on a non-uniform grid using the non-uniform Fast Fourier Transform (NUFFT)~\citep{fessler2003nonuniform}, producing a grid that has uniform frequency and phase coordinates. To better capture phase information, we represent complex Fourier coefficients with amplitudes and phase components. Absolute positional encoding and channel encoding are applied, and the tokens are then processed by a typical transformer network~\citep{vaswani2017attention}.

We evaluate \method's effectiveness on a variety of experiments. In the more common case of data that follows the Fourier power law, we evaluate Fourier-space natural image classification, MRI disease classification, and cryo-EM conformation classification and 3D reconstruction. To evaluate \method for the opposite case when data that does not conform to the power law, we test 2D crystal lattice identification, for which spectral energy is concentrated at higher frequencies~\citep{torquato2003local, torquato2018hyperuniform}. Across these settings \method consistently outperforms standard ViT and MLP baselines, with clear improvements on tasks requiring fine-grained discrimination under high noise. Our experiments verify the effectiveness of \method over different data domains and Fourier structures, showing the importance of non-uniform tokenization for capturing Fourier space information.

\section{Methods}
\label{sec:methods}

 
\subsection{Motivation}
\label{sec:background}

Data in Fourier space displays a wide variety of structures which are ill-captured by traditional ViT tokenization. As a motivating and important example, the variance of many natural signals is heavily concentrated in low spatial frequencies (Figure~\ref{fig:motivation}). With the standard discrete Fourier transform (DFT) convention that places the origin at the center of the Fourier image, most of the data variation is concentrated around the center. Additionally, the variance of many natural signals scales as a power law in frequency, such that low-frequency variances are orders of magnitude greater than high-frequency variances~\citep{field1987relations,van1996modelling}.

\begin{figure}
    \centering
    \includegraphics[width=\linewidth]{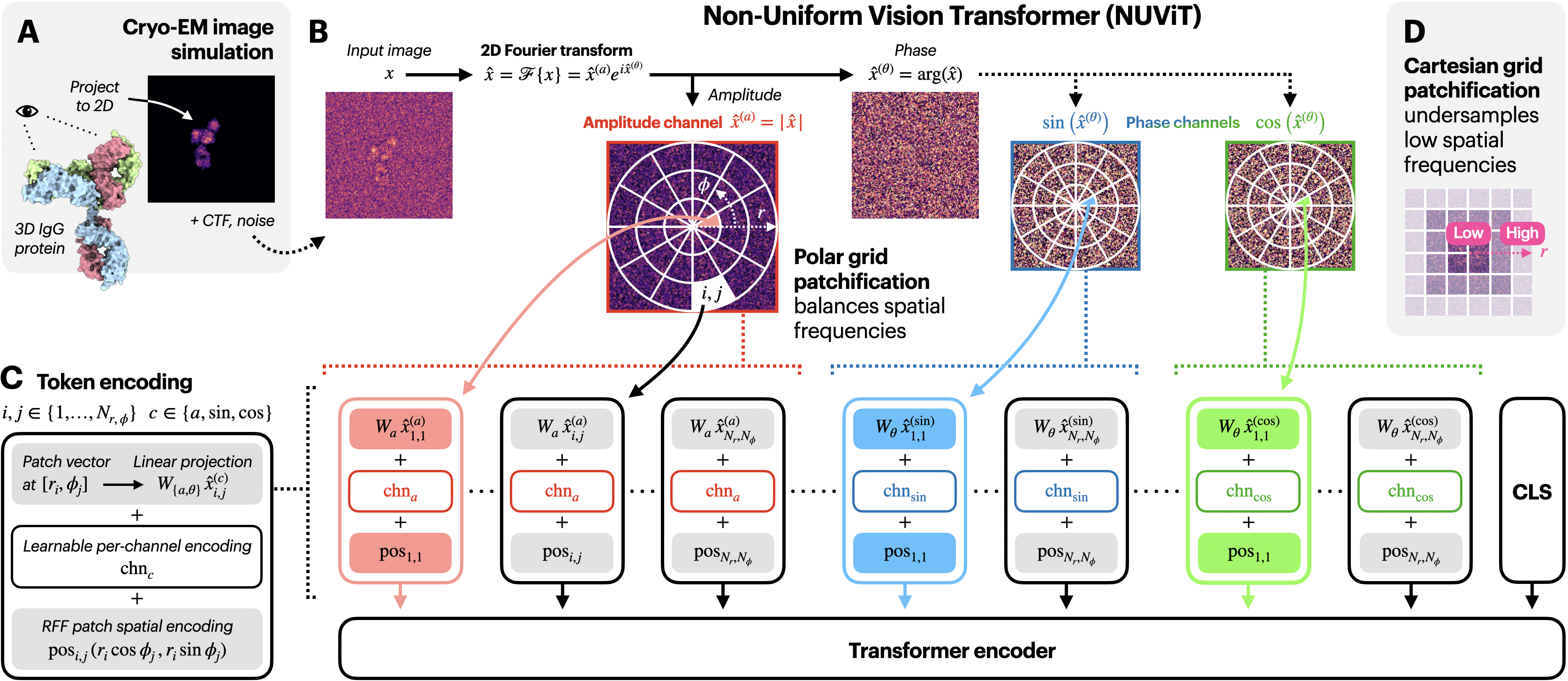}
    \vspace{-12pt}
    \caption{\textbf{NUViT architecture.} \textbf{A)} The cryo-EM image formation model is used to generate synthetic benchmark datasets. \textbf{B)} NUViT operates on Fourier-transformed 2D images using the NUFFT, where the data is represented in polar coordinates with amplitude $a$ and phase $\theta$  ($\sin$ and $\cos$) channels. Each channel map is partitioned into radial-angular patches and tokenized. \textbf{C)} Patches are encoded with channel and spatial position features. \textbf{D)} Traditional Cartesian grid patches undersample the high-variance low-frequency center of the Fourier-transformed data.}
    \label{fig:architecture}
    \vspace{-6pt}
\end{figure}

Standard ViT tokenization partitions an image into square patches on a uniform Cartesian grid. In Fourier space, this is poorly justified, since each square patch spans a mix of spatial frequencies rather than aligning with concentric frequency bands from the spectrum. Moreover, for many data modalities, this effectively undersamples the information-dense low-frequency components of a Fourier-space signal and oversamples the high-frequency periphery. NUViT addresses both issues by tokenizing on a non-uniform polar grid that allocates more samples to the low-frequency center (Figure~\ref{fig:architecture}).


 
\subsection{NUViT}
\label{sec:nuvit}

\subsubsection{Tokenization}
 
To address the sampling mismatch described above, we represent the Fourier data on a polar grid with $N_r$ radial shells and $N_\phi$ angular samples per shell, yielding coordinates $\{(r_i, \phi_j) : r_i = i / N_r,\, \phi_j = \pi j / N_\phi\}$ for $i = 1, \ldots, N_r$ and $j = 1, \ldots, N_\phi$. We use the NUFFT to evaluate the Fourier transform at these non-uniform coordinates, producing a $N_r \times N_\phi$ grid in polar space. This grid is treated as a regular 2D image and partitioned into square patches of size $P \times P$ using the standard ViT patching procedure. The total number of tokens is thus $N = (N_r / P) \times (N_\phi / P)$.
 
\subsubsection{Data Representation}
\label{sec:datarep}
 
Fourier-space data $\hat{x}$ lacks a canonical representation for deep learning. Fourier data $\hat{x}$ can be represented naively as complex numbers $\hat{x} \in \mathbb{C}^{D}$, as two channels of real and imaginary data $\hat{x} \in \mathbb{R}^{2D}$, or as amplitudes $a = |\hat{x}| \in \mathbb{R}^{D}$ and phases $\theta=\arg(\hat{x})\in [0, 2\pi)$. Since amplitudes follow a power law, log-amplitudes are also reasonable. If $x$ is real-valued, one can alternatively use the Hartley transform, which produces a real-valued output with no information loss.
 
In NUViT, we use an amplitude-and-phase representation. For each complex value $\hat{x}_j = a_j e^{i \theta_j}$, we construct three channels: the raw amplitude and a circular encoding of the phase. Concretely, the representation is:
\begin{align}
    \mathbf{r}(\hat{x}) = \bigl[\,[a_1, \ldots, a_D],\; [\cos(\theta_1), \ldots, \cos(\theta_D)],\; [\sin(\theta_1), \ldots, \sin(\theta_D)]\,\bigr]  \label{eq:datarep}
\end{align}
 
On a polar grid, each radial shell corresponds to a band of frequencies at a fixed magnitude. The amplitude-and-phase representation aligns naturally with this radial structure: the amplitude channel captures the power-law decay across shells, while the phase channels encode the within-shell variation. 
 
\subsubsection{Positional Encoding}
\label{sec:posenc}
 
NUViT uses two types of additive positional encoding: a spatial positional encoding that denotes the 2D coordinates of each patch, and a per-channel positional encoding that identifies the amplitude and phase channels.
 
\paragraph{Spatial positional encoding}
We encode the 2D center coordinate $\mathbf{x}_{ij} = (r_i, \phi_j)$ of each patch using random Fourier features (RFF)~\citep{tancik2020fourier}:
\begin{align}
    \gamma_{\mathrm{spatial}}(\mathbf{x}) = [\cos(2\pi \mathbf{B} \mathbf{x}),\; \sin(2\pi \mathbf{B} \mathbf{x})]^T, \quad \mathbf{B} \sim \mathcal{N}(0, \sigma^2 \mathbf{I})  \label{eq:rff}
\end{align}
where $\mathbf{B} \in \mathbb{R}^{d_{\mathrm{pe}} \times 2}$ is a matrix of random frequencies sampled once at initialization and held fixed during training, and $\sigma$ is a hyperparameter controlling the frequency bandwidth. We use RFF rather than the learned 1D embeddings of a standard ViT~\citep{dosovitskiy2020image} because the polar grid coordinates are inherently 2D and continuous-valued, making them a poor fit for a lookup table of discrete learned embeddings. RFF also allows the encoding to generalize to different grid resolutions at test time.
 
\paragraph{Channel positional encoding}
Our data representation produces three channels (amplitude, $\cos\theta$, $\sin\theta$) that have very different value ranges and semantic meanings. To preserve this channel-specific information, we follow~\citet{bao2023channel} and associate each channel $c \in \{a, \{\cos, \sin\}\}$ with a learnable embedding $\mathbf{e}_c \in \mathbb{R}^{D}$, which is added to the token embedding after projection. 
 
\subsubsection{Forward Pass}
\label{sec:forward}
 
Given an input image $x$, we first compute its 2D Fourier transform and evaluate it on the polar grid using the NUFFT, then construct the three-channel representation $\mathbf{r}(\hat{x})$ (Equation~\ref{eq:datarep}). The resulting $N_r \times N_\phi \times 3$ tensor is partitioned into $N = (N_r / P) \times (N_\phi / P)$ patches. Let $\mathbf{p}^{(c)}_k$ denote the $k$-th patch of channel $c$, flattened to a vector in $\mathbb{R}^{P^2}$.
 
Each patch is linearly projected to the transformer hidden dimension $D$, and both spatial and channel positional encodings are added:
\begin{align}
    \mathbf{z}^{(c)}_{k} = \mathbf{W}^{(c)}_{\mathrm{proj}} \, \mathbf{p}^{(c)}_k + \mathbf{b}+ \gamma_{\mathrm{spatial}}(\mathbf{x}_k) + \mathbf{e}_c  \label{eq:token}
\end{align}
where $\mathbf{W}^{(c)}_{\mathrm{proj}} \in \mathbb{R}^{D \times P^2}$ is a per-channel linear projection, $\mathbf{b}$ is a bias term, $\gamma_{\mathrm{spatial}}(\mathbf{x}_k)$ is the RFF encoding of the patch center coordinate (Equation~\ref{eq:rff}), and $\mathbf{e}_c$ is the learnable channel embedding. 
 
A learnable \texttt{[CLS]} token $\mathbf{z}_{\mathrm{cls}} \in \mathbb{R}^{D}$ is prepended to the sequence. The full input sequence is:
\begin{align}
    \mathbf{z}_0 = [\mathbf{z}_{\mathrm{cls}};\; \mathbf{z}^{(a)}_{1},\, \ldots,\, \mathbf{z}^{(a)}_{N},\; \mathbf{z}^{(\theta)}_{1},\, \ldots,\, \mathbf{z}^{(\theta)}_{N}]
    \label{eq:sequence}
\end{align}
This sequence of $2N + 1$ tokens is then processed by a standard transformer encoder~\citep{vaswani2017attention} consisting of $L$ layers of alternating multi-head self-attention (MSA) and MLP blocks with pre-norm layer normalization~\citep{dosovitskiy2020image}:
\begin{align}
    \mathbf{z}'_\ell &= \mathrm{MSA}(\mathrm{LN}(\mathbf{z}_{\ell-1})) + \mathbf{z}_{\ell-1}, \quad \ell = 1, \ldots, L  \label{eq:msa} \\
    \mathbf{z}_\ell &= \mathrm{MLP}(\mathrm{LN}(\mathbf{z}'_\ell)) + \mathbf{z}'_\ell  \label{eq:mlp}
\end{align}
The output representation is taken from the \texttt{[CLS]} token: $\mathbf{y} = \mathrm{LN}(\mathbf{z}^{0}_{L})$. For classification, a linear head is applied to $\mathbf{y}$. For cryo-EM reconstruction, $\mathbf{y}$ replaces the MLP encoder in cryoDRGN~\citep{zhongreconstructing, zhong2021cryodrgn} to produce a latent embedding per input image, which models the image class and conditions a 3D neural field for reconstruction.
 
\paragraph{Inductive bias}

Compared to a standard ViT, \method introduces additional inductive biases. An important one for machine learning is the assumption that the low spatial frequencies of the input data carry information content that is better sampled with a polar grid. This assumption is appropriate whenever the data obeys the Fourier power law, as is the case for natural images, MRI images, cryo-EM images, and many other modalities. The use of a polar grid also introduces a bias towards isotropy, which is common in many types of Fourier data, including Fourier power law data. The polar grid also respects the \emph{spectral autoregression} structure of Fourier data: if we group pixels into Fourier shells $[g_1, g_2, \ldots]$ by distance from the origin, the subsequences $[g_1], [g_1, g_2], \ldots$ encode the image at monotonically increasing resolution. Each ring of patches on the polar grid corresponds naturally to a frequency shell, so the model can attend to coarse-to-fine information in a structured way.



\section{Results}

We evaluate NUViT in five settings. To validate our design choices against simpler baselines, we test (1) Fourier-domain natural image classification and (2) 2D crystal lattice identification. To demonstrate NUViT's utility for different imaging modalities, we test (3) supervised MRI classification, (4) supervised cryo-EM image classification, and (5) heterogeneous cryo-EM reconstruction, which tests whether NUViT can serve as an effective encoder in a downstream generative model.

\subsection{Natural Image Classification}
\label{sec:img-cls}

\paragraph{Setup}
We first use Fourier-domain classification of natural images as a simple setting to evaluate NUViT and ablate its design choices. Input RGB images are converted to grayscale and Fourier transformed. We evaluate on CIFAR-10 and CIFAR-100~\citep{krizhevsky2009learning} using top-1 accuracy. We measure robustness by performing test-time inference on CIFAR-10-C and CIFAR-100-C~\citep{hendrycks2019benchmarking}. As Fourier-space baselines, we use a FIT~\citep{buchholz2022fourier}-derived model and a ViT-S~\citep{dosovitskiy2020image} with a Hartley transform representation. 

\paragraph{Results}
NUViT achieves the best overall performance on both datasets (Table~\ref{tab:img-cls}). FIT performs poorly, likely due to its one-token-per-pixel design, which produces extremely long sequences even for modestly sized images, leading to slow convergence.
We additionally ablate NUViT's components on the CIFAR task (Table~\ref{tab:ablations}). The polar grid is the single most important design choice ($-$25.14 points on CIFAR-10), followed by the data representation, where the amplitude and phase representation is most effective, though we note it requires more parameters. Using a channel positional encoding has a small impact on performance.


\begin{table}[t]
  \centering
  \begin{minipage}{0.5\textwidth}
    \centering
    \footnotesize
    \caption{\textbf{Natural image classification in Fourier space}. Top-1 test accuracy on CIFAR-10/100 and their corrupted counterparts CIFAR-10-C/100-C, evaluated using models trained only on the corresponding clean datasets.}
    \vspace{6pt}
    {
        \begin{tabular}{lcccc}
          \toprule
          & \multicolumn{4}{c}{\textbf{CIFAR}} \\
          \cmidrule(l){2-5}
          \textbf{Method} & \textbf{10} & \textbf{10-C} & \textbf{100} & \textbf{100-C} \\
          \midrule
          FIT & 23.35 & 16.55 & 5.21 & 2.76 \\
          ViT-S (Hartley) & 53.76 & 49.84 & 23.25 & \textbf{20.13} \\
          \textbf{NUViT-S (Ours)} & \textbf{68.63} & \textbf{52.99} & \textbf{27.20} & 18.69 \\
          \bottomrule
        \end{tabular}
      }
      \label{tab:img-cls}
  \end{minipage}
  \vspace{-6pt}
  \hfill
  \begin{minipage}{0.42\textwidth}
    \centering
    \footnotesize
    \caption{\textbf{Synthetic 2D crystal lattice classification.} Top-1 test accuracy. All encoders achieve 100\% train accuracy.}
    \vspace{6pt}
    {
        \begin{tabular}{lc}
          \toprule
          \textbf{Method} & \textbf{Crystal} \\
          \midrule
          Random chance & 25.0 \\
          MLP (real) & 39.8 \\
          MLP (Hartley) & 33.3 \\
          ViT-S (real) & 66.0 \\
          ViT-S (Hartley) & 58.7 \\
          \midrule
          \textbf{NUViT (Ours)} & \textbf{97.5} \\
          \bottomrule
        \end{tabular}
      }
      \label{tab:crystal-cls}
  \end{minipage}
  \vspace{-6pt}
\end{table}

\begin{table}[t]
  \centering
  \footnotesize
  \caption{\textbf{NUViT ablation study}. Key architectural components of NUViT are evaluated by top-1 classification accuracy on Fourier-transformed natural images.}
  \vspace{6pt}
  {
      \begin{tabular}{lcc}
        \toprule
        \textbf{Ablation} & \textbf{CIFAR-10} & \textbf{CIFAR-100}  \\
        \midrule
        Cartesian grid & 43.49 & 14.95 \\
        Real + Imaginary & 49.73 & 18.79 \\
        Hartley & 46.04 & 18.12 \\
        No channel positional encoding & 66.04 & \textbf{27.33} \\
        \textbf{NUViT (Ours)} & \textbf{68.63} & 27.20  \\
        \bottomrule
      \end{tabular}
    }
    \label{tab:ablations}
    \vspace{-6pt}
\end{table}

\subsection{2D Crystal Lattice Classification}


\paragraph{Setup} The domains discussed so far follow the Fourier power law with a negative exponent. To test whether \method generalizes to the opposite case, we benchmark on hyperuniform patterns, whose power spectra rise with frequency because they suppress fluctuations at large spatial scales~\citep{torquato2003local} — as observed in cosmological superhomogeneity~\citep{gabrielli2002glass} and crystal lattices. While perfect crystals do not follow a power law, imperfect crystals with random perturbations approximately follow $S(k) \propto k^\alpha$ with $\alpha = 2$~\citep{gabrielli2004point, kim2018effect, klatt2020cloaking}. We generate a dataset of such imperfect crystals with four types of symmetry, randomly displacing each lattice point (Figure~\ref{fig:crystal-examples}). As Fourier-space baselines, we use an MLP~\citep{zhongreconstructing} and a ViT-S~\citep{dosovitskiy2020image}, each with real and Hartley transform representations.

\paragraph{Results} In Table~\ref{tab:crystal-cls}, we find that \method substantially outperforms alternative encoders on 2D crystal lattice classification. Because every encoder reaches 100\% train accuracy, this gap reflects generalization rather than capacity. Each lattice class can be uniquely identified by its Bragg peaks, which are a set of peaks in the power spectrum of each crystal. However, identification is complicated by random perturbations, which dampen the peaks by the Debye-Waller factor $e^{-\sigma^2k^2}$, where $\sigma$ is the noise and $k$ is the frequency~\citep{ashcroft1976solid}. Therefore, Bragg peaks at higher frequency are more dampened than those at lower frequency. The higher test accuracy of \method compared to baseline methods provides evidence that our polar grid tokenization better matches the geometry of the spectrum than standard ViTs, capturing subtle peak information.

\subsection{MRI Classification}
\label{sec:mri}


\paragraph{Setup} To evaluate \method on MRI data, we use the OASIS Alzheimer's Detection Multi-Class dataset~\citep{shreyan_mohanty_2025}, which contains 2D brain MRI images labeled by Alzheimer's disease severity. To increase realism, we add MRI-specific corruptions and noise at a realistic signal-to-noise ratio (SNR), adapted from fastMRI~\citep{zbontar2018fastmri}. The task is challenging because it requires fine-grained distinctions between adjacent severity levels. Because these labels are ordinal, each encoder is paired with a CORAL head and trained with a class-weighted CORAL loss~\citep{cao2020rank}. We evaluate models with quadratic weighted kappa (QWK), which measures agreement with the ground truth relative to chance while penalizing errors in proportion to the squared distance between predicted and true labels.

\begin{table}[t]
\centering
\caption{\textbf{MRI classification.} Test QWK ($\uparrow$) for different image representations.}
\vspace{6pt}
\label{tab:mri}
\small
\begin{tabular}{lc}
\toprule
Method & Test QWK ($\uparrow$) \\
\midrule
ViT (real) & 0.266 \\
ViT (Hartley) & 0.232 \\
\textbf{NUViT (Ours)} & \textbf{0.348} \\
\bottomrule
\end{tabular}
\vspace{-6pt}
\end{table}


\paragraph{Results} Under realistic MRI conditions, \method achieves a meaningful gain in QWK over alternative encoders (Table~\ref{tab:mri}). Distinguishing adjacent severity levels requires detecting subtle anatomical differences in noisy images. \method's improvement provides evidence that polar grid tokenization is better at extracting information from the low signal-to-noise ratio (SNR) high frequency bands. Because MRI data is acquired directly in the frequency domain, these results suggest that \method could operate on raw MRI data without first reconstructing an image, which we leave to future work.

\subsection{Cryo-EM Image Classification}
\label{sec:conf-cls}

\paragraph{Setup}
We next evaluate NUViT on cryo-EM heterogeneity classification, where the goal is to classify noisy 2D projection images of dynamic proteins by their underlying 3D state. Cryo-EM images are characterized by extremely low signal-to-noise ratios and are commonly represented in the Fourier domain, making this a natural application for NUViT. We compare against two baselines corresponding to encoders used in cryo-EM reconstruction: the cryoDRGN~\citep{zhongreconstructing, zhong2021cryodrgn} MLP encoder and a ViT-S~\citep{dosovitskiy2020image} encoder, each tested with both real-space and Hartley transform representations. Hartley transform values are the default for many cryo-EM models~\citep{zhongreconstructing, zhong2021cryodrgn, zhong2021cryodrgn2, levy2022amortized, gu2026cryohype}. We do not test FIT~\citep{buchholz2022fourier} due to its computational cost and slow convergence.

We evaluate on one experimental and four synthetic datasets with a 90/10 train-test split. The synthetic datasets are from CryoBench~\citep{jeon2024cryobench}: \texttt{IgG-1D} and \texttt{IgG-RL} contain 100 conformational states of the IgG antibody, while \texttt{Ribosembly} and \texttt{Tomotwin-100} contain 16 and 100 distinct biomolecular species, respectively. The experimental dataset is \texttt{EMPIAR-10076}~\citep{davis2016modular}, containing 5 major states of the assembling 50S ribosome.

\begin{table}[b]
  \vspace{-10pt}
  \centering
  \footnotesize
  \caption{\textbf{Cryo-EM image classification.} This supervised classification task assigns labels to heterogeneous protein images. The datasets contain projection images of a compositional mixture of distinct structures (\texttt{Tomotwin-100}, \texttt{Ribosembly}, \texttt{EMPIAR-10076}) or conformational variations from the shape deformation of a single structure (\texttt{IgG-1D}, \texttt{IgG-RL})~\citep{jeon2024cryobench, davis2016modular}. \texttt{EMPIAR-10076} is a published experimental dataset, while the others are simulated. Performance is measured by top-1 accuracy. Results are reported as \texttt{train / test accuracy}.}
  \vspace{6pt}
  {
      \setlength{\tabcolsep}{3pt}
      \begin{tabular}{lccccc}
        \toprule
        \textbf{Method} & \textbf{IgG-1D} & \textbf{IgG-RL} & \textbf{Ribosembly} & \textbf{Tomotwin-100} & \textbf{EMPIAR-10076} \\
        \midrule
        Random chance & 1.00 & 1.00 & 6.25 & 1.00 & 6.25 \\
        MLP (real) & 99.33 / 1.25 & 99.60 / 2.10 & 99.61 / 42.40 & 99.50 / 42.12 & 99.65 / 39.94  \\
        MLP (Hartley) & 99.10 / 1.43 & 99.06 / 2.61 & 99.58 / 44.58 & 99.18 / 32.71 & 99.77 / 38.55 \\
        ViT-S (real) & 100.00 / 1.05 & 100.00 / 1.05 & 97.55 / 45.49 & 99.73 / 75.70 & 98.98 / 46.47 \\
        ViT-S (Hartley) & 1.66 / 1.41 & 57.82 / 8.11 & 98.01 / 72.56 & 98.84 / 84.32 & 98.73 / 50.89 \\
        \midrule
        \textbf{NUViT (Ours)} & 36.54 / \textbf{13.30} & 94.77 / \textbf{66.79} & 98.80 / \textbf{80.62} & 99.36 / \textbf{88.86} & 96.84 / \textbf{51.52} \\
        \bottomrule
      \end{tabular}
    }
    \label{tab:conf-cls}
    \vspace{-6pt}
\end{table}

\begin{figure}[t]
    \centering
    \includegraphics[width=\linewidth]{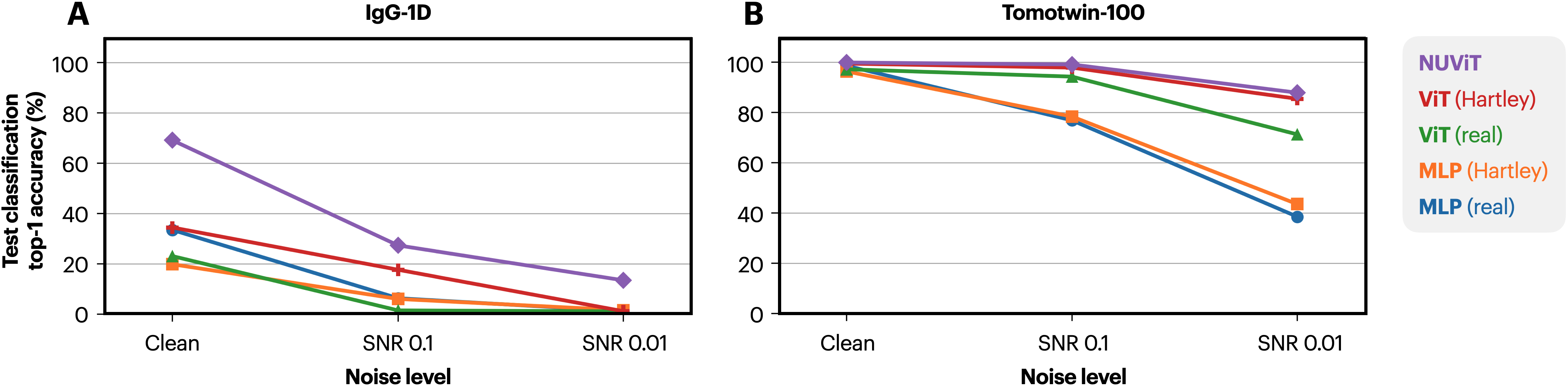}
    \vspace{-12pt}
    \caption{\textbf{Cryo-EM image classification noise robustness}. Comparison of test classification top-1 accuracy at different noise levels for simulated datasets \textbf{(A)} \texttt{IgG-1D} and \textbf{(B)} \texttt{Tomotwin-100}.}
    \label{fig:robustness}
    \vspace{-15pt}
\end{figure}

\paragraph{Results}
\method outperforms all Fourier-space baselines on heterogeneity classification (Table~\ref{tab:conf-cls}). The improvements are largest on tasks requiring fine-grained discrimination. On \texttt{IgG-1D} and \texttt{IgG-RL}, which require distinguishing small differences in the angle of a flexible linker region, all baseline encoders achieve near-random accuracy (1--8\% on 100 classes), whereas NUViT reaches 13.30\% and 66.79\%, respectively. While 13.30\% on \texttt{IgG-1D} is still low in absolute terms, it represents a qualitative shift: \method is the only encoder to extract meaningful conformational signal from these images. On \texttt{Ribosembly}, which requires distinguishing more complete from less complete ribosome assemblies, \method achieves 80.62\% compared to 72.56\% for the best baseline. On \texttt{Tomotwin-100}, a coarser-grained compositional classification task, \method achieves 88.86\% compared to 84.32\% for ViT-S (Hartley), while all other methods performed worse. On the experimental dataset \texttt{EMPIAR-10076}, \method achieves a small gain.

We also examine robustness to noise (Figure~\ref{fig:robustness}). High noise heavily degrades the performance of existing encoders, with the MLP encoder failing almost entirely at a signal-to-noise ratio (SNR) of 0.01. \method is substantially more robust to noise, likely because the polar grid allocates more capacity to low-frequency components, which have higher SNR than high-frequency components.

\subsection{Cryo-EM 3D Reconstruction}
\label{sec:cryoem-recon}

        

\begin{table}[b]
  \vspace{-10pt}
  \centering
  \footnotesize
  \caption{\textbf{Cryo-EM reconstruction.} Effect of encoder architecture on cryoDRGN~\citep{zhongreconstructing, zhong2021cryodrgn} reconstruction quality, measured by $\mathrm{FSC}_\mathrm{AUC} (\uparrow)$. All methods use the same decoder; only the encoder differs. $\mathrm{FSC}_\mathrm{AUC}$ statistics are calculated over the reconstructed volumes, and do not represent multiple model runs.}
  {
      \begin{tabular}{lcccccc}
        \toprule
        \multirow{2}{*}{\textbf{Encoder}} & \multicolumn{2}{c}{\textbf{IgG-1D}} & \multicolumn{2}{c}{\textbf{Tomotwin-100}} & \multicolumn{2}{c}{\textbf{EMPIAR-10076}} \\
        \cmidrule(lr){2-3} \cmidrule(lr){4-5} \cmidrule(lr){6-7}
        & Mean (std) & Median & Mean (std) & Med & Mean (std) & Med  \\
        \midrule
        MLP (Hartley) & 0.350 (0.019) & 0.357 & \textbf{0.346 (0.033)} & \textbf{0.349} & 0.418 (0.027) & 0.422 \\
        ViT (Hartley) & 0.363 (0.006) & 0.365 & 0.337 (0.027) & 0.340 & 0.423 (0.021) & 0.397 \\
        \textbf{NUViT (Ours)} & \textbf{0.367 (0.003)} & \textbf{0.368} & \textbf{0.345 (0.039)} & \textbf{0.348} & \textbf{0.433 (0.020)} & \textbf{0.435} \\
        \bottomrule
      \end{tabular}
    }
    \label{tab:cryoem_recon}
    \vspace{-6pt}
\end{table}

The goal of heterogeneous cryo-EM reconstruction is to recover multiple 3D structures from a dataset of noisy 2D projection images. Variational autoencoder~\citep{kingma2013auto} (VAE)-based methods such as cryoDRGN~\citep{zhongreconstructing, zhong2021cryodrgn} use an encoder to map each Fourier-space 2D image to a latent code, which a decoder uses to generate the corresponding Fourier-space 3D volume.

\paragraph{Setup}
We next evaluate whether NUViT can serve as an effective encoder in a downstream generative model for cryo-EM reconstruction. Specifically, we replace the MLP encoder in the cryoDRGN~\citep{zhongreconstructing, zhong2021cryodrgn} model with either a ViT (Hartley) encoder or a NUViT encoder, keeping the neural field decoder and all other components fixed. Beyond standard reconstruction quality, we test whether NUViT can enable amortized inference in cryo-EM. We note that existing encoders in cryo-EM reconstruction do not generalize beyond their training set: they are vulnerable to memorization~\citep{levy2025cryodrgn} and are not robust to rotations or shifts~\citep{edelberg2023using}, with the extreme noise in cryo-EM images being a major contributing factor.

For these experiments, we evaluate on two synthetic datasets, \texttt{IgG-1D} and \texttt{Tomotwin-100}~\citep{jeon2024cryobench}, and the experimental \texttt{EMPIAR-10076} dataset of the assembling 50S ribosome~\citep{davis2016modular}. Reconstruction quality is measured by per-image $\mathrm{FSC}_\mathrm{AUC}$~\citep{jeon2024cryobench}. For amortization experiments, we use a 90/10 train-test split for \texttt{EMPIAR-10076}. For the synthetic \texttt{IgG-1D} and \texttt{Tomotwin-100} datasets, we generate a new test set of 100k images following the cryo-EM image formation model~\citep{jeon2024cryobench}. 

\begin{figure}[t]
    \centering
    \includegraphics[width=\linewidth]{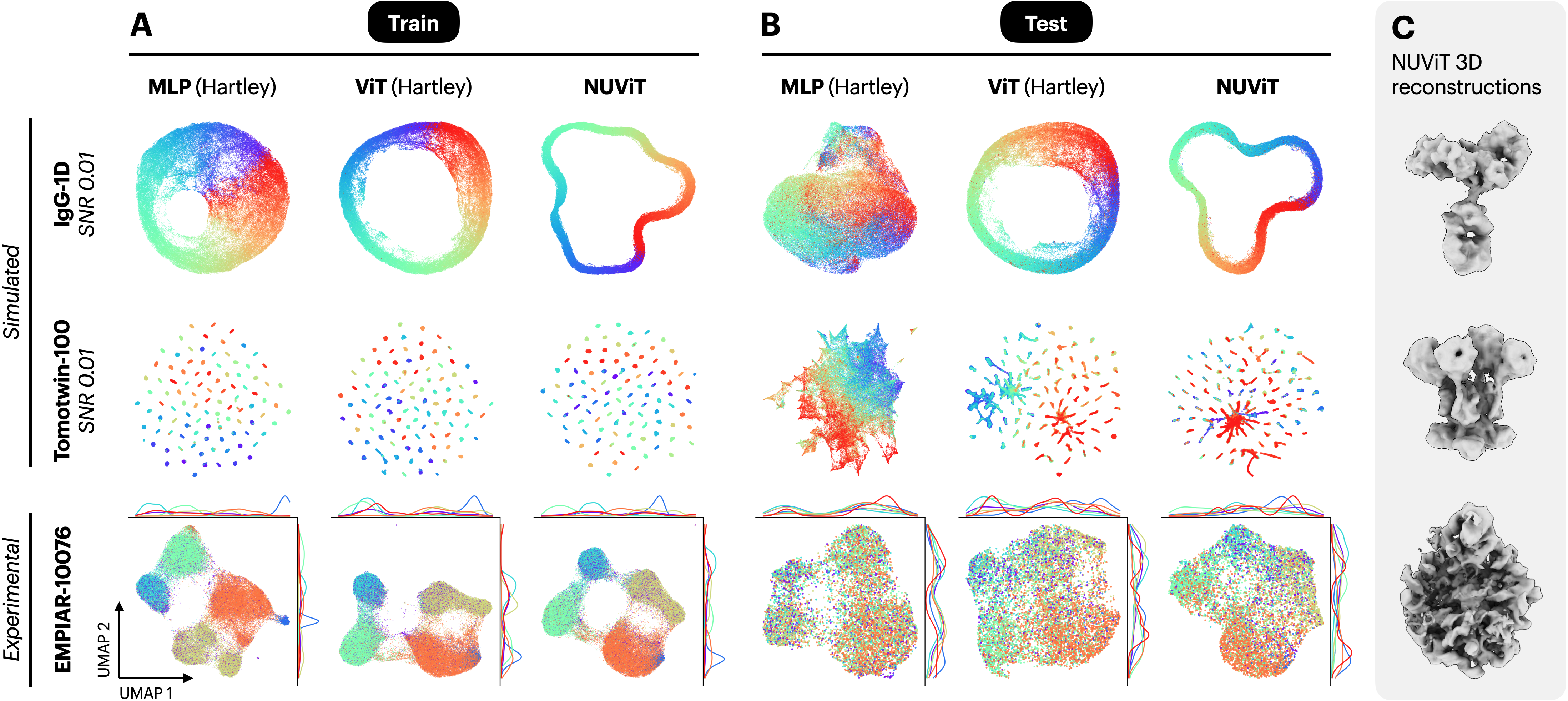}
    \vspace{-12pt}
    \caption{\textbf{Latent space visualizations of cryoDRGN with different Fourier image encoders.} \textbf{A,B)} UMAP~\citep{mcinnes2018umap} visualizations of train and test latent spaces. \texttt{IgG-1D} latent embeddings are colored by the ground truth Fab angle, \texttt{Tomotwin-100} is colored by the ground truth protein, and \texttt{EMPIAR-10076} is colored by the published major class labels~\citep{davis2016modular}. \textbf{C)} Example reconstructed density maps from our cryoDRGN+NUViT reconstruction model.}
    \label{fig:latentspace}
    \vspace{-15pt}
\end{figure}

\paragraph{Reconstruction quality}
Replacing the encoder with NUViT improves reconstruction quality across all three datasets (Table~\ref{tab:cryoem_recon}).
We compare against additional methods in Table~\ref{tab:cryoem_recon_si_iggtt}. On \texttt{IgG-1D}, we find that using cryoDRGN-NUViT achieves state-of-the-art performance among all deep learning methods, trailing only RECOVAR~\citep{gilles2025cryo}, a linear subspace method that is able to decode multi-views at inference time via adaptive kernel regression, giving it an advantage over cryoDRGN's decoder.

\paragraph{Amortization}
We next evaluate each encoder on held-out test images. These images are passed through the encoder to generate latent embeddings (Figure~\ref{fig:latentspace}), which are then passed through the decoder to reconstruct volumes that are compared with ground truth maps via $\mathrm{FSC}_\mathrm{AUC}$ (Table~\ref{tab:amortization}). The MLP encoder shows the largest train-test gap, with $\mathrm{FSC}_\mathrm{AUC}$ dropping from 0.346 to 0.139 on \texttt{Tomotwin-100}. ViT encoders improve amortization substantially, and NUViT improves further, achieving the best test performance on all three datasets. The latent space visualizations in Figure~\ref{fig:latentspace} are consistent with these findings: NUViT's latent space is better organized on both train and test images, whereas the MLP encoder fails to disentangle different conformations or compositions on test images. The improved amortization is plausibly because NUViT's polar grid tokenization concentrates capacity on the low-frequency components that carry the most structural information, making the encoder less susceptible to overfitting to the high-frequency noise that dominates individual training images.

        

\begin{table}[t]
  \vspace{-10pt}
  \centering
  \footnotesize
  \caption{\textbf{Amortization of cryo-EM reconstruction models.} We show the ability of different Fourier-space encoder architectures with cryoDRGN~\citep{zhongreconstructing} for reconstructing unseen cryo-EM images of the proteins the reconstruction model was trained on, measured by $\mathrm{FSC}_\mathrm{AUC}$. }
  {
      \begin{tabular}{lcccccc}
        \toprule
        \multirow{2}{*}{\textbf{Encoder Method}} & \multicolumn{2}{c}{\textbf{IgG-1D}} & \multicolumn{2}{c}{\textbf{Tomotwin-100}} & \multicolumn{2}{c}{\textbf{EMPIAR-10076}} \\
        \cmidrule(lr){2-3} \cmidrule(lr){4-5} \cmidrule(lr){6-7}
        & Mean (std) & Median & Mean (std) & Med & Mean (std) & Med \\
        \midrule
        MLP (Hartley) (original)  & 0.337 (0.018) & 0.336 & 0.139 (0.104) & 0.120 & 0.400 (0.041) & 0.413 \\
        ViT (Hartley) & 0.354 (0.016) & 0.359 & 0.293 (0.101) & 0.331 & 0.399 (0.035) & 0.411 \\
        \textbf{NUViT (Ours)} & \textbf{0.361 (0.011)} & \textbf{0.365} & \textbf{0.301 (0.105)} & \textbf{0.341} & \textbf{0.409 (0.037)} & \textbf{0.422} \\
        \bottomrule
      \end{tabular}
    }
    \label{tab:amortization}
    \vspace{-15pt}
\end{table}

\section{Related Work}

\paragraph{Fourier transforms in deep learning}
Fourier transforms have seen a variety of uses in deep learning, including as a data representation for modalities including audio, medical imaging, and microscopy, a way to do efficient convolution~\citep{gu2021efficiently, fu2022hungry} or token mixing~\citep{rao2021global, guibas2021adaptive, lee2022fnet, nguyen2022fourierformer}, positional encodings~\citep{tancik2020fourier}, a shortcut for efficient 3D reconstruction via the Fourier Slice Theorem~\citep{zhong2021cryodrgn}, and more. However, architectures designed for Fourier-space data remain underexplored, especially for transformer architectures~\citep{vaswani2017attention, dosovitskiy2020image}. Fourier Image Transformers (FIT)~\citep{buchholz2022fourier} considers Fourier-space data arising from tomographic reconstruction. FIT proposes to unroll a Fourier-space image into a 1D sequence of pixels ordered by frequency, with one token per pixel. FIT focuses on tomographic reconstruction and super resolution and does not apply a polar grid or NUFFT for building a general-purpose architecture. \citep{alshaabi2025fourier} proposes a two-stage Fourier transformer for aberration correction, but retains the Cartesian ViT~\citep{dosovitskiy2020image} patching scheme. While not specifically designed for Fourier-space data, another line of work focuses on designing architectures for complex-valued data~\citep{trabelsi2017deep, venkatasubramanian2024steinmetz}, which could also be applied to the frequency domain. However, these methods suffer from challenges in optimization, lack robust activation functions, and are not well-validated for frequency-space data.   

\paragraph{Non-uniform grids for Fourier space in deep learning}
In deep learning, the use of non-uniform grids for Fourier space data has been mainly explored in the context of Fourier Neural Operators (FNOs)~\citep{li2020fourier}, which are resolution-invariant networks designed to solve PDEs. FNOs use repeated blocks of residual global convolutions which consist of a Fourier transform, linear transform, and inverse Fourier transform. FNOs do not use all frequencies, and in practice the high frequency modes are truncated, leaving only the low frequency modes. This approach is feasible because many PDEs follow a Fourier power law. Since FNOs implement the Fourier transform as an FFT, they are limited to Cartesian grids. \cite{lingsch2023beyond} examines the case of arbitrary grids, and shows that FNOs on irregular grids computed using the NUFFT can outperform FNOs on equispaced Cartesian grids. Our method extends this to general transformer architectures.

\section{Conclusion and Limitations}

We propose \textbf{Non-Uniform Vision Transformers (NUViT)}, a customized transformer architecture for Fourier space data. Motivated by the Fourier power law, NUViT introduces the idea of sampling and tokenizing on a non-uniform polar grid, providing stronger inductive biases such as better sampling of important low-frequency regions of Fourier space that are undersampled by a traditional ViT. Extensive experiments show that NUViT is a more expressive, generalizable, and robust encoder for Fourier data. Despite these encouraging results, NUViT has the limitation that it is more computationally expensive than ViTs. This may be addressed by using more performant NUFFT libraries, such as the one proposed in~\citep{lingsch2023beyond}, as well as reducing the oversampling rate in the NUFFT~\citep{beatty2005rapid}. 
This work could have long-term societal benefits through improved image reconstruction and analysis techniques for applications like cryo-EM and MRI, contributing to deeper insight into biological and physical systems.

\subsubsection*{Acknowledgements}

The authors would like to thank Ariana-Dalia Vlad, Joseph Clark, Alkin Kaz, Minkyu Jeon, Ziyu Xiong, and Rishwanth Raghu for helpful discussions. The authors acknowledge the use of computing resources at Princeton Research Computing, a consortium of groups led by the Princeton Institute for Computational Science and Engineering (PICSciE) and Office of Information Technology's Research Computing. The Zhong lab is supported by the Chan Zuckerberg Initiative DAF (2025-358484), an advised fund of Silicon Valley Community Foundation, the National Institutes of Health (DP2GM164606), the AI2050 program at Schmidt Sciences (G-25-69788),  Janssen Pharmaceuticals, Generate Biomedicines, the Princeton Catalysis Initiative, and the Princeton School of Engineering and Applied Sciences. The funders had no role in study design, data collection and analysis, decision to publish or preparation of the manuscript.



\bibliography{iclr2027_conference}
\bibliographystyle{iclr2027_conference}

\newpage
\appendix

\setcounter{table}{0}
\renewcommand{\thetable}{S\arabic{table}}
\setcounter{figure}{0}
\renewcommand{\thefigure}{S\arabic{figure}}

\section{Training details}
\label{sec:training}

\subsection{Implementation Details}
Unless otherwise stated, all results are for the NUViT-S (Small) configuration, which is analogous to the ViT-S: $L = 12$ layers, hidden dimension $D = 384$, $12$ attention heads, and MLP dimension $4D = 1536$. For the polar grid, we use $H = W = 128$ with patch size $P = 16$, yielding $N = 64$ patches per channel and $2 \times 64 + 1 = 129$ tokens. We set the RFF bandwidth to $\sigma = 1$. We also find it useful to control the max radius of the radial shells, denoted $R$. Unless otherwise stated, all NUViT models process polar grids of size $N_r = N_\phi = 128$. All our experiments are done in PyTorch~\citep{paszke2019pytorch}. We implement NUFFTs using the \texttt{TorchKbNUFFT} package~\citep{muckley:20:tah}. 

\subsection{Compute Details}
Models were trained on a shared research cluster. Most models were trained with a single NVIDIA H100 GPU, 32GB of memory, and 32 CPUs. When training on larger datasets such as \texttt{Ribosembly}, we use 128GB of memory.


\begin{table}[htbp]
\centering
\caption{\textbf{Computational cost.} Wall-clock time per training epoch and peak memory usage.}
\vspace{6pt}
\label{tab:compute-cost}
\small
\begin{tabular}{lrrrr}
\toprule
\textbf{Encoder} & \textbf{Params} & \textbf{Time / epoch} & \textbf{GPU Mem.} & \textbf{CPU Mem.} \\
\midrule
MLP   & 19.9\,M & 2:19.62 & 12.3\,GB & 7.9\,GB \\
ViT   & 24.9\,M & 1:40.42 & 13.6\,GB & 7.9\,GB \\
NUViT & 25.1\,M & 2:11.84 & 16.7\,GB & 7.8\,GB \\
\bottomrule
\end{tabular}
\vspace{-6pt}
\end{table}

In Table~\ref{tab:compute-cost}, we provide the wall-clock time and memory for a cryo-EM reconstruction model trained on the ADH dataset, using an NVIDIA H100 GPU. NUViT takes ~31\% more wall-clock time per epoch and ~23\% more memory than a ViT. Compared to an MLP, NUViT is slightly faster in wall clock time but uses more GPU memory. All encoders have similar CPU memory usage. We believe that there are still untapped potential gains in efficiency for NUViT encoders, including using a more efficient implementation, such as that of~\citep{lingsch2023beyond}. Although the speed of calculating one NUFFT is fast, reducing the oversampling rate during NUViT from 2x to 1.25x could provide further speedups.

\subsection{Natural Image Classification}

For the natural image experiments we process the data as follows:
\begin{enumerate}
    \item First, we resize all images to $128 \times 128$ resolution so we can use the same model for all experiments.
    \item We apply \texttt{RandomCrop}, \texttt{RandomHorizontalFlip}, and \texttt{ColorJitter} data augmentations and normalize the data. 
    \item We transform the images from RGB to grayscale, apply the Fourier transform, and put the Fourier transformed-data into the desired data representation. 
\end{enumerate}
All models were trained with batch size 64 with the Adam optimizer~\citep{kingma2014adam} with learning rate 1e-4 for 100 epochs. For \method on CIFAR-10, we use a log amplitude and phase representation and set $R$ to 2. On CIFAR-100, we use an amplitude and phase representation and set $R$ to 4. Both models were trained with a cosine learning rate schedule. \method takes approximately 18 hours to train. Hyperparameters were chosen by grid search. 

\subsection{MRI Classification}

The OASIS Alzheimer's Detection Multiclass Dataset~\citep{shreyan_mohanty_2025} consists of the brain MRI scans of 416 subjects. The dataset is designed to be data-leakage free, so any single patient's data is contained in a single training split. Alzheimer's severity was assessed based on the Clinical Dementia Rating (CDR) into four distinct classes: non-demented (CDR 0.0), very mild demented (0.5), mild demented (1.0), and moderate demented (2.0). The OASIS dataset is unbalanced towards non-demented patients, with a class breakdown of 316/70/28/2. 

\subsection{2D Crystal Lattice Classification}

\begin{figure}[t]
    \centering
    \includegraphics[width=\linewidth]{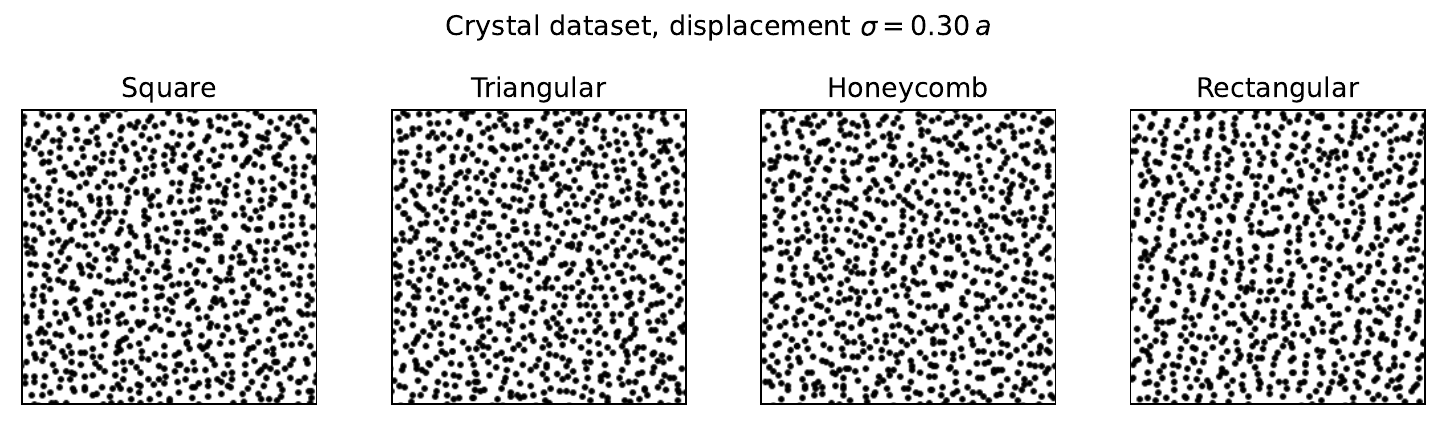}
    \vspace{-12pt}
    \caption{\textbf{Simulated 2D crystal lattice examples.}}
    \label{fig:crystal-examples}
    \vspace{-6pt}
\end{figure}

The synthetic 2D crystal lattice dataset consists of four classes of crystal lattices (square, triangular, honeycomb, rectangular) with 1000 train examples and 250 test examples per class, for a total of 4000 training examples and 1000 test examples. Imperfect lattices were generated by randomly perturbing each point by 30\% of the lattice spacing (Figure~\ref{fig:crystal-examples}). The perturbation value was calibrated on ViT-S (real) to ensure the dataset was neither too easy nor too difficult, with the final value being 30\% of the lattice spacing. Each model was trained for 100 epochs without data augmentation. Model-specific hyperparameters were chosen based on the best parameters from natural image classification.

\subsection{Cryo-EM Image Classification}

For cryo-EM heterogeneity classification, we use a random 90/10 train/test split for each of the datasets. All models were trained with batch size 64 with the Adam optimizer~\citep{kingma2014adam} with learning rate 1e-4. We use $R$ values of $8, 16, 16, 16, 16, 4, 8$ for \texttt{IgG-1D}, \texttt{IgG-RL}, \texttt{Ribosembly}, \texttt{Tomotwin-100}, \texttt{EMPIAR-10076}, \texttt{RyR}, respectively, and $N_\phi = 128$ for all datasets. 

\subsection{Cryo-EM 3D Reconstruction}
For the synthetic datasets (\texttt{IgG-1D} and \texttt{Tomotwin-100}), all models were traind on the whole dataset. To assess amortization, we generated an additional 100K images based on CryoBench~\citep{jeon2024cryobench}. Models were trained on 20, 50, and 100 epochs for \texttt{IgG-1D}, \texttt{EMPIAR-10076}, and \texttt{Tomotwin-100}, respectively. Otherwise, cryoDRGN-MLP used cryoDRGN's default parameters were used. CryoDRGN-\method used the best model parameters from the previous section. 

\section{Metrics}

Following CryoBench~\citep{jeon2024cryobench}, we use per-image FSC to jointly evaluate reconstruction quality and heterogeneity. In cryo-EM, the Fourier Shell Correlation (FSC) curve is a standard tool
for comparing two volumes by measuring their correlation across concentric shells in Fourier space. Since cryo-EM methods typically reconstruct a 3D maps from a single 2D image, per-
image FSC quantifies how well each individual reconstruction matches the groundtruth. $\mathrm{FSC}_\mathrm{AUC}$ summarizes performance using the area under the FSC curve. To capture the distribution of reconstruction quality across conformations, we sample
one image per conformation, as in CryoBench. The maximum value of $\mathrm{FSC}_\mathrm{AUC}$ is 0.5.

\section{Additional results}

\subsection{Cryo-EM Image Classification}


\begin{table}[htbp]
\centering
\caption{\textbf{Cryo-EM image classification (top-5 accuracy).} Performance is measured by top-5 accuracy. Results are reported as \texttt{train / test accuracy}. }
\vspace{6pt}
\label{tab:conf-cls-top5}
\small
\begin{tabular}{lcc}
\toprule
\textbf{Method} & \textbf{IgG-1D} & \textbf{IgG-RL} \\
\midrule
Random chance & 5.00 / 5.00 & 5.00 / 5.00 \\
MLP (real) & 99.99 / 5.57 & 100.00 / 9.08 \\
MLP (Hartley) & 99.99 / 6.59 & 99.98 / 11.54 \\
ViT-S (real) & 100.00 / 4.77 & 100.00 / 5.84 \\
ViT-S (Hartley) & 7.96 / 7.19 & 92.30 / 28.54 \\
\midrule
\textbf{NUViT (Ours)} & 86.63 / \textbf{59.01} & 99.95 / \textbf{96.78} \\
\bottomrule
\end{tabular}
\vspace{-6pt}
\end{table}

In Table~\ref{tab:conf-cls-top5}, we show top-5 accuracy results for cryo-EM conformation classification on the IgG-1D and IgG-RL datasets, providing context where most results are close to random chance. These datasets are challenging because they require the discrimination of very similar conformations at different viewing angles and in the presence of high noise, with IgG-1D being more difficult since each conformation of IgG-1D has a more similar neighboring conformation than IgG-RL. We find that most baselines for IgG-1D perform near random chance, due to the difficulty of the classification problem. For IgG-RL, most methods perform significantly better than random chance, unlike for top-1 accuracy, indicating some discrimination ability. As with top-1 accuracy, for both datasets our method significantly outperforms the baselines. 

\subsection{Cryo-EM 3D Reconstruction}


\begin{table}[htbp]
\centering
\caption{\textbf{Additional experimental datasets.} $\mathrm{FSC}_\mathrm{AUC} (\uparrow)$ by training dataset from EMPIAR-10028~\citep{wong2014cryoem} and CAHRA~\citep{feathers2026cahra}.}
\vspace{6pt}
\label{tab:si-addl-exp}
\small
\begin{tabular}{lccc}
\toprule
\textbf{Dataset} & \textbf{MLP} & \textbf{ViT} & \textbf{NUViT (Ours)} \\
\midrule
EMPIAR-10028 & 0.471 & 0.478 & \textbf{0.480} \\
CAHRA-ADH & 0.390 & 0.448 & \textbf{0.481} \\
CAHRA-Aldolase & 0.363 & 0.437 & \textbf{0.477} \\
CAHRA-PKM2 & 0.402 & 0.448 & \textbf{0.462} \\
CAHRA-ADH/Aldolase/PKM2 & 0.437 & 0.448 & \textbf{0.455} \\
\bottomrule
\end{tabular}
\vspace{-6pt}
\end{table}

\paragraph{Additional experimental datasets} Table~\ref{tab:si-addl-exp} shows the performance of different encoders for cryo-EM reconstruction on the EMPIAR-10028~\citep{wong2014cryoem} and subsets of the CAHRA Heterogeneity Challenge~\citep{feathers2026cahra} datasets. The ADH, Aldolase, PKM2 label refers to a dataset composed of a mixture of the three proteins. We find that \method encoders outperform ViT encoders, which in turn outperform MLP encoders.

\begin{table}[htbp]
  \centering
  \small
  \caption{\textbf{Cryo-EM reconstruction on IgG-1D and Tomotwin-100}. We show the ability of NUViT to serve as a Fourier image encoder for a VAE-based cryo-EM reconstruction model (cryoDRGN~\citep{zhongreconstructing, zhong2021cryodrgn}). Performance is measured by per-image area under the Fourier shell correlation curves ($\mathrm{FSC}_\mathrm{AUC} (\uparrow)$). }
  \vspace{6pt}
  {
      \begin{tabular}{lcccc}
        \toprule
        \multirow{2}{*}{\textbf{Method}} & \multicolumn{2}{c}{\textbf{IgG-1D}} & \multicolumn{2}{c}{\textbf{Tomotwin-100}} \\
        \cmidrule(lr){2-3} \cmidrule(lr){4-5}
        & Mean (std) & Median & Mean (std) & Med \\
        \midrule
        3D classification~\citep{punjani2017cryosparc} & 0.297 (0.019) & 0.291 & 0.046 (0.026) & 0.037 \\
        3DVA~\citep{punjani20213d} & 0.349 (0.004) & 0.350 & 0.088 (0.040) & 0.077 \\
        RECOVAR~\citep{gilles2025cryo} & 0.386 (0.005) & 0.388 & 0.258 (0.109) & 0.254 \\
        \midrule
        CryoDRGN~\citep{zhongreconstructing} & 0.350 (0.019) & 0.357 & 0.346 (0.033) & 0.349 \\
        CryoDRGN-AI~\citep{levy2025cryodrgn} & 0.364 (0.002) & 0.364 & 0.202 (0.044) & 0.321 \\
        OPUS-DSD~\citep{luo2023opus} & 0.335 (0.026) & 0.339 & 0.237 (0.049) & 0.251 \\
        3DFlex~\citep{punjani20233dflex} & 0.335 (0.003) & 0.335 & - & - \\
        SFBP~\citep{kimanius2022sparse} & - & - & 0.036 (0.011) & 0.036 \\
        CryoHype~\citep{gu2026cryohype} & - & - & 0.346 (0.033) & 0.353 \\
        \midrule
        CryoDRGN + ViT (Hartley) & 0.363 (0.006) & 0.365 & 0.337 (0.027) & 0.340 \\
        \textbf{CryoDRGN + NUViT (Ours)} & \textbf{0.367 (0.003)} & \textbf{0.368} & 0.345 (0.039) & 0.348 \\
        \bottomrule
      \end{tabular}
    }
    \label{tab:cryoem_recon_si_iggtt}
    \vspace{-6pt}
\end{table}


\paragraph{Expanded comparison to prior methods} Table~\ref{tab:cryoem_recon_si_iggtt} shows the performance of our method compared to state-of-the-art deep learning and non-deep learning models. On \texttt{IgG-1D}, NUViT achieves state-of-the-art performance among deep learning methods, trailing only RECOVAR~\citep{gilles2025cryo}, a PCA-based model which uses a more powerful adaptive kernel regression decoder. On Tomotwin-100, NUViT reaches the performance of the best methods. All baseline numbers, except those for CryoDRGN, were drawn from CryoBench~\citep{jeon2024cryobench} and CryoHype~\citep{gu2026cryohype}. Dashes indicate that there is no prior evaluation of that baseline on the given dataset.

\paragraph{Latent space quality}

To assess the latent spaces of Figure~\ref{fig:latentspace} quantitatively, we use two sets of metrics. To evaluate continuous heterogeneity with a known groundtruth coordinate structure, such as IgG-1D, we evaluate trustworthiness~\citep{venna2001neighborhood} and continuity~\citep{venna2006local}, two neighborhood-preservation metrics designed to assess continuous latent spaces such as IgG-1D's ring latent space. Both metrics take values in $[0, 1]$ with higher values being better. We compute these metrics between a latent space and an idealized ring structure. 

\begin{table}[htbp]
\centering
\caption{\textbf{Embedding quality.} Trustworthiness and continuity ($\uparrow$) of the learned embedding space.}
\vspace{6pt}
\label{tab:igg1d-trust-cont}
\small
\begin{tabular}{lcccc}
\toprule
 & \multicolumn{2}{c}{\textbf{Train}} & \multicolumn{2}{c}{\textbf{Test}} \\
\cmidrule(lr){2-3} \cmidrule(lr){4-5}
\textbf{Encoder} & Trust. & Cont. & Trust. & Cont. \\
\midrule
MLP   & 0.9260 & 0.8467 & 0.7534 & 0.6910 \\
ViT   & 0.9454 & 0.9429 & 0.8791 & 0.8910 \\
\textbf{NUViT (Ours)} & \textbf{0.9634} & \textbf{0.9692} & \textbf{0.9111} & \textbf{0.9323} \\
\bottomrule
\end{tabular}
\vspace{-6pt}
\end{table}

In Table~\ref{tab:igg1d-trust-cont}, we find that \method's train and test latent space is much closer to the ideal ring structure of IgG-1D's groundtruth reaction coordinate than the latent spaces corresponding to the other encoders. We believe this is also reflected visually by \method's latent space being a nice, thin loop, whereas other latent spaces look thicker and noisier. 

Datasets exhibiting significant compositional heterogeneity, such as Tomotwin-100, ideally form many discrete clusters in latent space. To assess these types of datasets, we use Adjusted Rand Index (ARI) and Adjusted Mutual Information (AMI), which are commonly used to assess these datasets in cryo-EM.  

\begin{table}[htbp]
\centering
\caption{\textbf{Clustering quality.} Adjusted Rand Index (ARI) and Adjusted Mutual Information (AMI), $\uparrow$ higher is better.}
\vspace{6pt}
\label{tab:tt100-ari-ami}
\small
\begin{tabular}{l cc cc}
\toprule
 & \multicolumn{2}{c}{\textbf{Train}} & \multicolumn{2}{c}{\textbf{Test}} \\
\cmidrule(lr){2-3} \cmidrule(lr){4-5}
\textbf{Encoder} & ARI & AMI & ARI & AMI \\
\midrule
MLP   & \textbf{0.9952} & \textbf{0.9968} & 0.1356 & 0.3601 \\
ViT   & 0.9892 & 0.9944 & 0.6924 & 0.8194 \\
\textbf{NUViT (Ours)} & 0.9838 & 0.9942 & \textbf{0.7067} & \textbf{0.8222} \\
\bottomrule
\end{tabular}
\vspace{-6pt}
\end{table}

In Table~\ref{tab:tt100-ari-ami}, we find that although all encoders perform extremely well on the training data, the latent space of the MLP encoder collapses when performing inference on test data, which matches the non-clustered latent space we observe in Figure~\ref{fig:latentspace}. \method again has the best-performing latent space, the ViT encoder's performance is only slightly less. Since one of the main drivers of reconstruction error is assigning an image the incorrect conformation (i.e., the wrong latent), this partially explains the reconstruction results of Tables~\ref{tab:cryoem_recon} and~\ref{tab:amortization}.

\subsection{Ablations}

\begin{table}[htbp]
\centering
\caption{\textbf{Ablation studies.} Top-1 accuracy (\%) on Tomotwin-100 and IgG-1D.}
\vspace{6pt}
\label{tab:si-ablations}
\small
\begin{tabular}{lcc}
\toprule
\textbf{Method} & \textbf{Tomotwin-100} & \textbf{IgG-1D} \\
\midrule
\multicolumn{3}{l}{\textit{Channel positional encoding}} \\
No channel PE & \textbf{90.45} & 7.62 \\
Channel PE    & 88.86 & \textbf{12.84} \\
\midrule
\multicolumn{3}{l}{\textit{Feature representation}} \\
Hartley          & 67.68 & 4.91 \\
Real/Imaginary   & \textbf{90.01} & 11.04 \\
Amplitude/Phase  & 88.86 & \textbf{13.30} \\
\bottomrule
\end{tabular}
\vspace{-6pt}
\end{table}

In Table~\ref{tab:si-ablations}, we repeat the ablation study of Table~\ref{tab:ablations} for cryo-EM conformation classification on the IgG-1D and Tomomtwin-100 datasets. We find that while channel PE and an amplitude-phase representation may not always be the optimal choice, they perform better on difficult datasets such as IgG-1D. As such, our default method may represent a strong starting point while various features may need to be tuned for optimal performance on any single dataset.  

\end{document}